\documentclass[letterpaper]{article} 
\usepackage{aaai24}  
\usepackage{times}  
\usepackage{helvet}  
\usepackage{courier}  
\usepackage[hyphens]{url}  
\usepackage{graphicx} 
\usepackage{natbib}  
\usepackage{caption} 
\usepackage{algorithm}
\usepackage{algorithmic}

\usepackage{amsfonts}
\usepackage{bbding}

\usepackage{newfloat}
\usepackage{listings}
\DeclareCaptionStyle{ruled}{labelfont=normalfont,labelsep=colon,strut=off} 
\floatstyle{ruled}
\newfloat{listing}{tb}{lst}{}
\floatname{listing}{Listing}
\title{EPCL: Frozen CLIP Transformer is An Efficient Point Cloud Encoder}
\author{
	Xiaoshui Huang\textsuperscript{\rm 1}\equalcontrib, Zhou Huang\textsuperscript{\rm 2}\equalcontrib, Sheng Li\textsuperscript{\rm 2}\equalcontrib, Wentao Qu\textsuperscript{\rm 2}\\
	Tong He\textsuperscript{\rm 1}\thanks{Corresponding author.}, Yuenan Hou\textsuperscript{\rm 1}, Yifan Zuo\textsuperscript{\rm 2}\footnotemark[2], Wanli Ouyang\textsuperscript{\rm 1}
}
\affiliations{
	\textsuperscript{\rm 1}Shanghai AI Lab, \textsuperscript{\rm 2}Jiangxi University of Finance and Economics 
	
}

\usepackage{bibentry}

\begin{document}

\maketitle

\begin{abstract}
The pretrain-finetune paradigm has achieved great success in NLP and 2D image fields because of the high-quality representation ability and transferability of their pretrained models. However, pretraining such a strong model is difficult in the 3D point cloud field due to the limited amount of point cloud sequences. This paper introduces \textbf{E}fficient \textbf{P}oint \textbf{C}loud \textbf{L}earning (EPCL), an effective and efficient point cloud learner for directly training high-quality point cloud models with a frozen CLIP transformer. Our EPCL connects the 2D and 3D modalities by semantically aligning the image features and point cloud features without paired 2D-3D data.  Specifically, the input point cloud is divided into a series of local patches, which are converted to token embeddings by the designed point cloud tokenizer. These token embeddings are concatenated with a task token and fed into the frozen CLIP transformer to learn point cloud representation. The intuition is that the proposed point cloud tokenizer projects the input point cloud into a unified token space that is similar to the 2D images.  Comprehensive experiments on 3D detection, semantic segmentation, classification and few-shot learning demonstrate that the CLIP transformer can serve as an efficient point cloud encoder and our method achieves promising performance on both indoor and outdoor benchmarks. In particular, performance gains brought by our EPCL are $\textbf{19.7}$ AP$_{50}$ on ScanNet V2 detection, $\textbf{4.4}$ mIoU on S3DIS segmentation and $\textbf{1.2}$ mIoU on SemanticKITTI segmentation compared to contemporary pretrained models. Code is available at \url{https://github.com/XiaoshuiHuang/EPCL}.
\end{abstract}

\section{Introduction}

Recently, the pretrain-finetune paradigm has achieved great success in natural language processing (NLP) \cite{chowdhery2022palm,gu2021ppt} and 2D image fields \cite{alayrac2022flamingo,dosovitskiy2020vit,radford2021learning}. In the pretrain-finetune paradigm, a backbone is first pretrained on a large-scale dataset to learn general and transferable representations. Then, the pretrained model is finetuned on training samples of the downstream task to learn task-specific knowledge. 

The CLIP models \cite{radford2021learning} are strong pretrained models, which are trained in the contrastive manner by leveraging more than 400 million image-text pairs. The impressive performance of CLIP models on few-shot and zero-shot tasks is attributed to the powerful representation learned from the large quantity of pretraining data and the inherent alignment between the image and language domains.

\begin{figure}
	\centering
	\includegraphics[width=\linewidth]{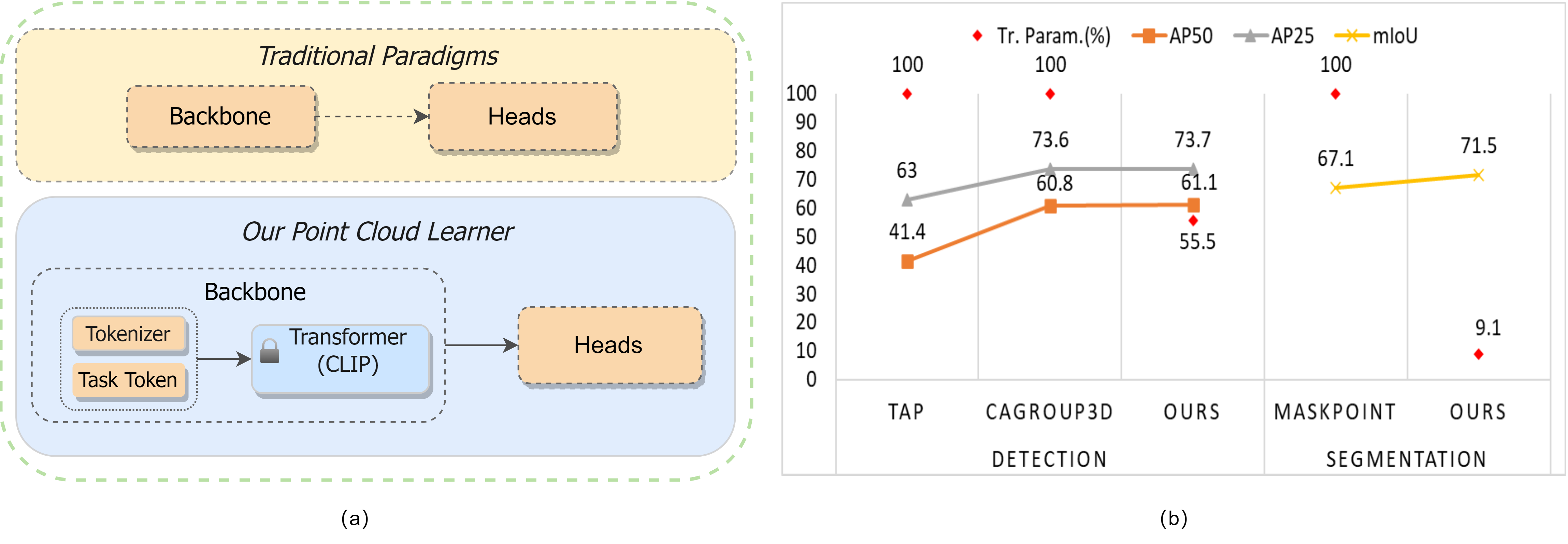}
	\caption{(a) Traditional paradigm fine-tunes the whole model, while our method only fine-tunes the tokenizer (T) and head (H). The CLIP transformer, which is initialized from the original CLIP weight, is kept frozen during training. (b) Our EPCL brings accuracy gains with higher training efficiency compared to SOTA pre-training methods.
	}
	\label{fig:f1}
\end{figure}

However, directly applying the pretrain-finetune paradigm to the point cloud field will confront great difficulties due to the scarcity of training samples as well as the inherent domain gap between point cloud and image domains. For instance, the majority of pretraining methods \cite{pang2022masked,qian2022pix4point,yu2022point,xie2020pointcontrast} are trained with limited data, ShapeNet \cite{shapenet2015} or ScanNet datasets \cite{dai2017scannet}. ShapeNet contains about 50, 000 objects and ScanNet contains 1, 513 room scans \cite{xie2020pointcontrast}. Compared to the pretraining data of CLIP, the number of training samples in the point cloud field is merely ten thousandth. The prior knowledge learned from the limited training samples is also limited.

Inspired by the great success of CLIP, we ask a question: \emph{can we apply the CLIP transformer to point cloud tasks as a pretrained encoder}? If the answer is yes, the 2D and 3D modalities are bridged and we can leverage the pretrained CLIP transformer for learning effective representations in the point cloud field. In this condition, the heavy reliance on 3D pre-training data can also be relieved.

To mitigate the domain gap between point cloud and image domains, we design an extremely efficient module, \emph{i.e.}, the point tokenizer, to map the point cloud and image information into the same embedding space. Since the point cloud and images all describe the surface information, we can consider them as a unified 2D-manifold that every point/pixel has a neighbourhood homeomorphic to a certain region of space $\mathbf{R}^2$ \cite{pressley2010elementary}. We hypothesise that the frozen CLIP can extract meaningful representation from the 2D-manifold input. The tokenizer embeds the point/pixel neighbourhoods into the unified token space and weakly aligns the features of 2D images and 3D point cloud. Then, the transformer encoder extracts meaningful representations by semantically aligning them further. We validate this hypothesis in the experimental results (Figure~\ref{fig:fig_corre}). 

Based on this finding, we propose the \textbf{E}fficient \textbf{P}oint \textbf{C}loud \textbf{L}earning (EPCL) framework to directly leverage the frozen CLIP transformer as the encoder  for point cloud tasks. The difference between our method and other 3D pre-training methods is illustrated in Figure \ref{fig:f1}. Our EPCL merely requires the training of the lightweight task token, tokenizer and task head while previous 3D pre-training algorithms need to the train the whole model. Take S3DIS segmentation as an example. The trainable parameters of our EPCL barely account for $9.1$\% of all trainable model parameters, which strongly demonstrates the superior efficiency of EPCL. 

\begin{figure}
	\centering
	\includegraphics[width=0.8\linewidth]{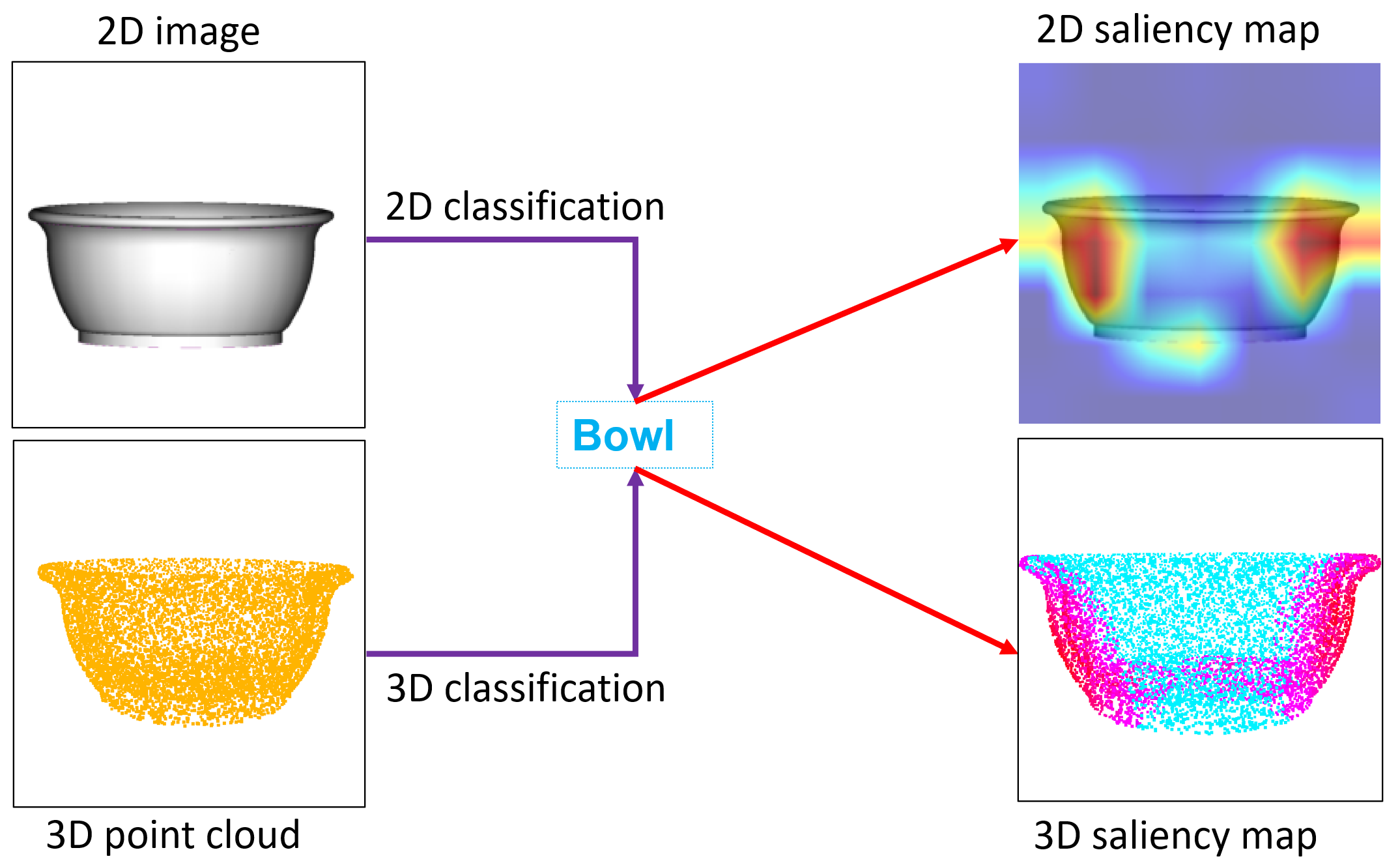}
	\caption{Using the frozen CLIP image transformer as an encoder for 2D and 3D classification, the saliency maps show the frozen CLIP model can attend to similar regions at different modalities.}
	\label{fig:f2}
\end{figure}

To sum up, our EPCL framework possesses the following merits: 

\textbf{High efficiency.} Our EPCL fully leverages the rich and broad knowledge hidden in CLIP models and only requires the training of a small portion of trainable network parameters. Our EPCL is apparently much more efficient compared to contemporary 3D pretraining algorithms that train all network parameters.

\textbf{Aligning 2D and 3D models without paired data}. Aligning multi-modal models has become mainstream to train a strong pretrained model while the existing methods \cite{radford2021learning,alayrac2022flamingo} usually require paired data, \emph{e.g.}, image-text pairs \cite{radford2021learning} and video-text pairs \cite{alayrac2022flamingo}. Our EPCL does not need 3D-2D paired data to train the model when adapting the CLIP image transformer for 3D tasks. Figure \ref{fig:f2} reveals that our method can semantically align similar regions in the 3D point cloud compared to CLIP in the 2D image on the recognition task.

\textbf{Free from 3D pre-training.} 
The strong 2D pre-trained CLIP transformer is directly used for 3D point clouds in EPCL without requiring 3D pre-training, which helps circumvent the barrier from the scarcity of 3D data.  

\textbf{Facilitating few-shot learning in downstream tasks.} Leveraging the rich knowledge learned in CLIP, EPCL is effective when downstream tasks have scarce training samples.

We perform comprehensive experiments on mainstream point cloud tasks including detection, segmentation, recognition, classification and few-shot learning. Experimental results show that EPCL achieves better performance than the state-of-the-art 3D pre-training methods. Notably, our EPCL brings the gains of $\textbf{19.7}$ AP$_{50}$ on ScanNet V2 detection, $\textbf{4.4}$ mIoU on S3DIS segmentation and $\textbf{1.2}$ mIoU on SemanticKITTI segmentation compared to contemporary pretrained encoders.

\textbf{Difference to prior works.}
While there have been several existing works proposing the utilization of 2D pretrained models, our method differs from them. For instance, Image2Point \cite{xu2022image2point} expands 2D kernels of a CNN into 3D kernels for point cloud feature extraction. Pix4Point \cite{qian2022pix4point} initializes from 2D pretrained backbones and finetunes the entire neural network, resulting in low training efficiency. PPKT \cite{liu2021learning} pretrains 3D backbones by distilling from 2D pretrained models, but it exhibits relatively low performance.
ACT \cite{dong2022autoencoders} adopts a two-stage strategy, training the teacher from a 2D pretrained model and distilling the teacher to a 3D point cloud Transformer student through masked modeling. However, prior works utilizing 2D pretrained models often suffer from either \emph{low performance} or \emph{low efficiency}.

In contrast to these prior works, our method, EPCL, \emph{directly applies} the frozen CLIP model to extract point cloud features for various tasks, achieving better performance compared to recent pretrained methods. In this GPT era, our approach provides an \emph{efficient} encoder for point cloud feature extraction. Additionally, since CLIP is a general vision pretrained model without task-specific information, we have designed a task token to further embed task-related biases.

\section{Related Work}

\subsection{CLIP-based methods}

CLIP~\cite{radford2021learning}, which aims to learn transferable visual representation from natural languages, has attracted increasing attention due to its promising results on various downstream tasks~\cite{lei2015predicting,kornblith2019better,recht2019imagenet}. It consists of two encoders for visual and text representations, respectively. The method is jointly trained to align the two modalities with over 400 million image-text pairs. The rich semantic representation shared by both domains inspires many works and has been demonstrated effective in tasks like image caption~\cite{vinyals2015show} and video~\cite{carreira2019short}. CLIPCAP~\cite{mokady2021clipcap} trained a lightweight mapping network to generate meaningful captions, while the CLIP and language model is frozen. EVL~\cite{lin2022frozen} addressed the task of zero-shot video understanding via contrastive learning between video and text representations, which is free from the annotation of the downstream tasks. CLIP-ViL \cite{shen2021much} uses CLIPs as pretrained backbones and finetunes the CLIP model on specific vision-language tasks.
LAMM \cite{yin2023lamm} uses frozen CLIP to embed multiple modalities into tokens and input these tokens into large language model to conduct multi-modal understanding tasks. Although promising, directly applying CLIP to 3D tasks is non-trivial due to the significant domain gap. 

\subsection{Point cloud representation learning}
Learning discriminative point cloud representation plays a critical role in downstream tasks. The existing methods can be divided into two categories, \emph{i.e.}, point-based and voxel-based methods.

\begin{figure*}
	\centering
	\includegraphics[width=\linewidth]{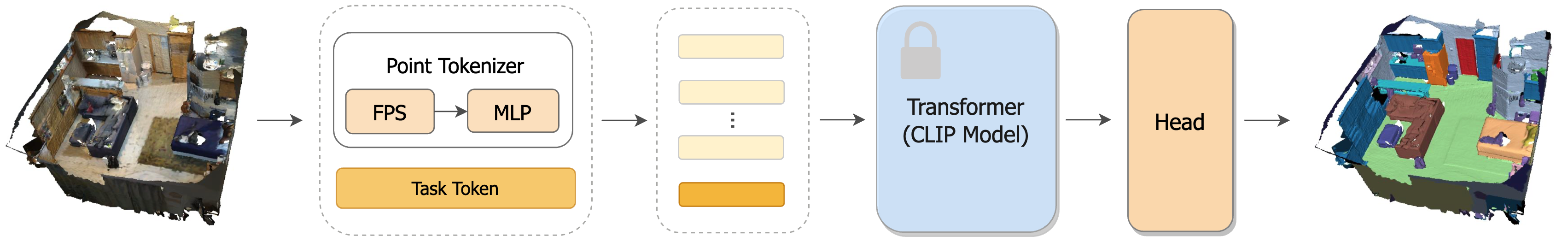}
	\caption{Schematic overview of EPCL. The Point Tokenizer contains two successive steps, that are Farthest Point Sampling (FPS) for downsampling the input point cloud and Multi-Layer Perceptron (MLP) for extracting features from the downsampled point cloud. The Task Token is task-specific and learnable. Tokens from the point tokenizer and task token are fed into the frozen CLIP Transformer. The Head uses the tokens from the Transformer to yield the predictions for each specific downstream task. The CLIP transformer, which is initialized from the original CLIP weight, is kept frozen during the training stage, while the point cloud tokenizer, task token and head are trainable.}
	\label{fig:f3}
\end{figure*}

\textbf{Point-based methods} extract discriminative representation from raw points using either multi-layer perception \cite{qi2017pointnet}, graph convolution \cite{wang2019dynamic} or kernel-based convolution \cite{thomas2019kpconv}. The objective of these methods is to leverage the global structure information or local property of point neighbours to describe the 3D point cloud.  The advantages of these methods are that features can directly extract from analyzing point neighbours, the memory consumption is relatively small and no preprocessing steps are required.

\textbf{Voxel-based methods} require to pre-process the given point clouds into voxels. Then, voxel-based convolution neural networks are applied to extract the representation. Typical examples are VoxelNet \cite{riegler2017octnet} and Minkowski Engine \cite{choy20194d}. They design octree-based convolution and sparse convolution to effectively extract the local representation of the point cloud without large GPU memory consumption. The advantage of these methods is that the representation can easily overcome the density variation.

\subsection{3D pre-training}
Pre-training aims to learn prior knowledge from the training data. The existing pre-training methods can be divided into three categories, \emph{i.e.}, global contrastive, local contrastive and Masking AutoEncoder (MAE). The global contrastive learning methods \cite{wang2021unsupervised,mei2022unsupervised} compare the global feature difference of point clouds. In contrast, local contrastive learning methods \cite{xie2020pointcontrast, wang2023take} compare the local point feature differences or local view pixel differences. Recently, the MAE strategy is applied to the point cloud field and several pre-training methods \cite{yu2022point,pang2022masked} are proposed to learn pretrained transformer backbones. These methods leverage the knowledge of pretrained datasets so that the downstream task models initialize from a better starting point.

Recently, several methods are proposed to use the 2D pre-trained models on point cloud tasks. PointCLIP \cite{zhang2022pointclip} projects the point cloud into 2D views and directly uses the frozen 2D pre-trained models for 3D recognition. P2P\cite{wang2022p2p} designs a projector to project the 3D objects into the 2D plane and designs several prompts to use the frozen 2D pre-trained backbones. PPKT \cite{liu2021learning}  transfers the knowledge of 2D pretraiend model to 3D backbones by using point-to-pixel loss.
However, these methods require projecting the 3D objects into several 2D views and are sensitive to view projection. Hence, they are used for object-level point clouds but face great difficulty in handling scene-level point cloud perception. 
Image2Point \cite{xu2022image2point} expands 2D kernels of a 2D CNN into 3D kernels and applied them to voxel-based point cloud tasks, which suffers from relatively low accuracy as the parameter domain gap. Pix4Point \cite{qian2022pix4point} initializes from 2D pretrained backbones and finetunes the whole neural network, which is not efficient. 
ACT \cite{dong2022autoencoders} requires two stage training to transfer the knowledge of 2D pretrained model to 3D point cloud transformer. However, the training process is not efficient and the performance has a large gap to task-specific model.

Different to previous approaches, our EPCL directly utilizes the pre-trained 2D CLIP transformer as an efficient encoder to extract point cloud features. And our method is applicable to both real-world and synthetic point cloud tasks. In this GPT era, our work provides the insights that frozen CLIP can achieve comparable or better performance to recent SOTA pretrained methods with higher efficiency.

\section{Method}
\label{sec:method}
This section first introduces the Vision Transformer (ViT)~\cite{dosovitskiy2020vit} in the 2D image field and the transformer in the point cloud field. Then, we present our EPCL and the rationale behind the workability of EPCL.

\subsection{Preliminary}
\textbf{2D Vision Transformer.} 
Given an image $I \in \mathbb{R}^{H\times W\times C}$, the ViT~\cite{dosovitskiy2020vit} divides the image into a sequence of flattened local image patches $x_p \in \mathbb{R}^{N\times {(P^2\cdot C)}}$ and uses a tokenizer to convert these patches into a 1D sequence of visual token embeddings $E_I(I)\in \mathbb{R}^{N\times D}$, where $N$ is the number of tokens, $P\times P$ is the image patch size, $D$ is the dimension of each image token. $H$ and $W$ are the height and width of the given image, respectively. The total number of patches is $N=HW/(P^2)$. The position embedding is concatenated to the visual token embeddings.  Visual tokens and class tokens are fed into the transformer for feature extraction. Afterwards, the feature is fed into the classification head to yield the classification results. Mathematically, the 2D ViT can be formulated as follows:
\begin{eqnarray}
	& z_0  = [x_{\mathrm{cls}}, E_I(I_{1,1}),...,
	E_I(I_{\frac{H}{P},\frac{W}{P}})] + E_{\mathrm{pos}},\\
	& \widetilde{z}_l  = \mathrm{MSA}(\mathrm{LN}(z_{l-1})+z_{l-1}), \\
	& z_l=\mathrm{MLP}(\mathrm{LN}(\widetilde{z}_l))+\widetilde{z}_l, \\
	& y=H^{\mathrm{cls}}(\mathrm{LN}(z_{L}^0)),
\end{eqnarray}
where $E_I(.)$ is the image tokenizer that extracts the token embedding for each image patch, and $x_{cls}$ is the class token.  The transformer consists of $L$ layers of layer normalization $LN(.)$, multi-head self-attention $\mathrm{MSA}(.)$ and multi-layer perceptron $\mathrm{MLP}(.)$. The residual connection is applied after every block. $H^{\mathrm{cls}}$ represents the classification head and takes the feature of the class token at the last layer as input. Take the 1000-class image classification task as an example. $H^{\mathrm{cls}}$ refers to a single MLP that maps the input feature into a 1000-dimension classification output.

\textbf{Transformer in point cloud.} Before the standard transformer is applied to the point cloud field, there are some transformer layers \cite{zhao2021point,guo2021pct} specifically designed for point cloud processing. Pioneered by PointBERT~\cite{yu2022point}, the standard transformer has been applied to point cloud tasks. Similar to the ViT, the point cloud $P\in \mathbb{R}^{A\times 3}$ is divided into a sequence of point cloud patches $P_p\in \mathbb{R}^{M\times (3K)}$, where $A$ and $M$ denote the number of points and patches, respectively, and $K$ denotes the number of points in each patch. These patches are sent to a tokenizer to extract point token embeddings. Then, these point token embeddings, position embedding and class token are fed into the standard transformer for feature extraction. Afterwards, these features are fed into a task head for downstream tasks. The Transformer for point cloud can be formulated as follows:
\begin{eqnarray}
	&  f_0  = [x_{\mathrm{cls}}, E_p(P_{1}),..., E_p(P_{M})] + E_{\mathrm{pos}},\\
	&  \widetilde{f}_l  = \mathrm{MSA}(\mathrm{LN}(f_{l-1})+f_{l-1}), \\
	&  f_l=\mathrm{MLP}(\mathrm{LN}(\widetilde{f}_l))+\widetilde{f}_l, \\
	&  y=H_p^{\mathrm{cls}}(\mathrm{LN}(f_{L}^0)),
\end{eqnarray}
where the $E_p$ is the point cloud tokenizer, the three-layer MLP is usually applied for obtaining point cloud token embeddings. $H^{cls}_p$ refers to three-layer MLPs that map the input feature into the $C$-dimension classification predictions for $C$-class point cloud classification tasks.

\textbf{Comparison between 2D and 3D transformers.}
By comparing the equation (1)-(3) and (5)-(8), the standard transformer module is the same, which consists of a series of the LN, MSA and MLP. The only difference lies in the tokenizer during the feature extraction. Then, the deep features are fed into different 2D/3D task heads for 2D/3D downstream tasks. Here, we want to investigate whether the same standard transformer module pretrained on 2D could be directly applied to 3D point cloud tasks.

\subsection{The proposed algorithm: EPCL}
The motivation of our method is to leverage the frozen 2D CLIP model for downstream point cloud understanding tasks. To this end, we propose the Efficient Point Cloud Learning (EPCL) framework to use the 2D frozen CLIP transformer and only finetune the tokenizer, task token and task head. The overall framework is shown in Figure \ref{fig:f3}. This section introduce the details of tokenizer, task token and Frozen CLIP transformer. The details of task head are attached in the supplement.

\paragraph{Point cloud tokenizer.}
Given a point cloud $P \in \mathbb{R}^{A\times 3}$, similar to the objective of the 2D tokenizer, the point cloud tokenizer aims to convert the input point cloud into a sequence of token embeddings. Specifically, we first sample $M$ points as centers of point patches by the Farthest Point Sampling (FPS) algorithm, and then group $K$ points from each center by the $K$-Nearest Neighbourhood (KNN) algorithm and thus obtain $M$ patches. These patches are further fed into several MLPs to obtain the token embeddings $E_p(P) \in \mathbb{R}^{M\times D_p},i\in[1...M]$, where $D_p$ is the dimension of each point token, \emph{i.e.}, 768.

\paragraph{Task token.} 
Since the CLIP is trained by a large-scale text-image pair dataset, it is lacked of task information. To further embed the given point clouds into a shared token space that benefit for the task, we design a task token to learn a global task-related bias. Our task token module is implemented by a fully connected layer with learnable parameters. Following~\cite{liu2021p}, we initialize the task token as enumerated numbers.

\paragraph{Frozen CLIP transformer.}
After the input point cloud is converted into a sequence of visual tokens, we feed visual tokens and task tokens into the CLIP image transformer, which is initialized from the original CLIP weight and kept frozen during training. The frozen CLIP transformer serves as the feature extractor for downstream point cloud tasks. 

\begin{figure}[]
	\centering
	\includegraphics[width=\linewidth]{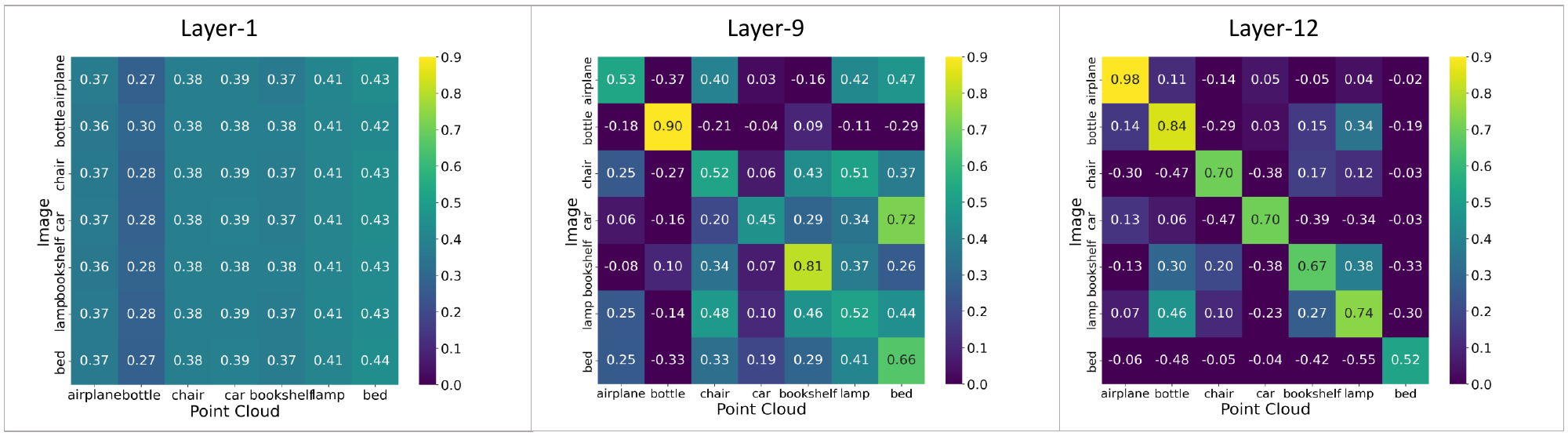}
	\caption{The cross-correlation between CLIP image features and point cloud features at layers 1, 9, and 12 for different object categories.}
	\label{fig:fig_corre}
\end{figure}

\subsubsection{Analysis on 2D-3D semantic alignment of CLIP transformer}
To analyse the workability of frozen 2D CLIP transformer for point cloud representation learning, we calculate the semantic similarity between image features and point cloud features. Specifically, we first calculate their feature cross-correlation at different layers of the same CLIP transformer. Then, we use the transformer explanation tool in~\cite{chefer2021generic} to obtain the significance map. For the 2D image, we crop a view from the ShapeNet model to keep the texture and apply the CLIP model to classify the image view. 

\textbf{Statistical results.} Figure \ref{fig:fig_corre} shows that the tokenizer can weakly align the 2D and 3D features. At shallow layers, the features from the point cloud and image for the same category have lower cross-correlation in the left sub-figure. As the layers go deeper, the 2D-3D features are matched with high cross-correlation in the same category (see the right sub-figure).  

\begin{figure}[ht]
	\centering
	\includegraphics[width=0.7\linewidth]{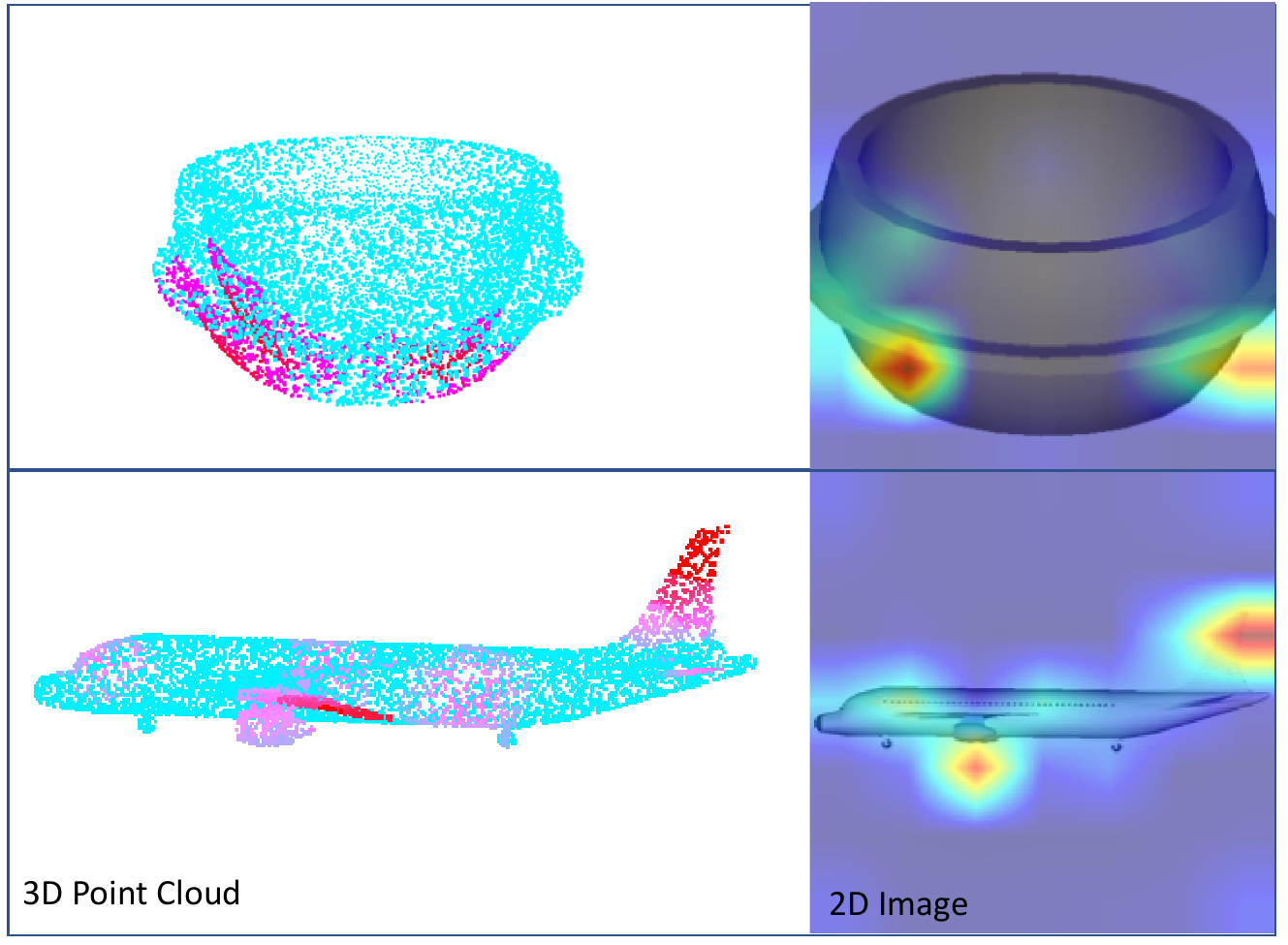}
	\caption{The semantic similarity between 2D image and 3D point cloud from significance maps.}
	\label{fig:fig_sim}
\end{figure}

\textbf{Visual results.} Figure \ref{fig:fig_sim} shows the roughly similar significance maps at a 2D image and 3D point cloud. This figure shows that the frozen CLIP model can capture similar semantic regions from 2D and 3D modalities. 

\subsubsection{The rationale behind EPCL}
To better understand why the frozen CLIP transformer is workable for the point cloud, we provide an intuitive explanation from the manifold aspect.
We define the input token space as $\Omega_I$ and output token space as $\Omega_O$. The CLIP image transformer learns a function $f$ to map the input tokens $X\in\Omega_I$ into semantically meaningful tokens $Y\in\Omega_O$: $Y=f(X)$. Since the CLIP has been trained in a large-scale dataset that contains diverse web image-text pairs, the input token space $\Omega_I$ is large and diverse.

The image tokenizer uses convolution to aggregate local information into the image token space, denoted by $\Omega_I^I$. Similarly, our point cloud tokenizer uses the FPS + KNN + MLP to aggregate local neighbourhood information token space, denoted by $\Omega_I^P$. Since a point cloud frame only records points on the surface, according to the manifold definition \cite{pressley2010elementary}, a small local point cloud patch is approximately a plane and the 3D point cloud lies in the 2D manifold. Since the given point cloud consists of many local planes, our tokenizer and the task token learn to project the 2D-manifold point cloud into the token space $\Omega_I^P$ that is similar to the CLIP image token space $\Omega_I^I$ projected from 2D image plane. 
Since the images and local point cloud patches are both 2D-manifold planes, the above tokenizer learning is achievable. Previous research shows that the transformer inherently extracts shape-biased features \cite{park2022vision,naseer2021intriguing} for 2D images. Therefore, the CLIP image transformer can extract shape-based features from the token space $\Omega_I^P$  for the given point cloud.

\section{Experiments}
\label{sec:experiments}
We introduce the details of used datasets and baselines in section \ref{sec:exp1}. Then, experiments of the downstream tasks are described in section \ref{sec:exp3},  section \ref{sec:exp4} and section \ref{sec:exp2}, respectively. Afterwards, the ablation studies are presented in section \ref{sec:exp5}.

\subsection{Datasets and baselines}
\label{sec:exp1}
\textbf{Datasets.} We conduct real-world detection on ScanNet \cite{dai2017scannet}, indoor semantic segmentation on S3DIS \cite{armeni20163d} and outdoor semantic segmentation on SemanticKITTI \citet{behley2019semantickitti}. Also, we evaluate the accuracy of few-shot learning and classification on synthetic ModelNet40 \cite{wu20153d}. 

\textbf{Baselines.}  Our EPCL focuses on leveraging pre-training for downstream tasks. For a fair comparison, the state-of-the-art (SOTA) point cloud pre-training methods with the transformer-based architectures are selected as the baselines. 
\begin{itemize}
	\item \textbf{Detection:} Following MaskPoint \cite{yu2022point}, we compare with MaskPoint \cite{yu2022point}, PointBERT\cite{pang2022masked}, TAP \cite{wang2023take}, Simple3D-Former \cite{wang2022can}, SoftGroup \cite{vu2022softgroup} and CAGroup3D \cite{wang2022cagroup3d}.
	\item \textbf{Segmentation:} Following Simple3D-former \cite{wang2022can}, we compare with MaskPoint \cite{pang2022masked} and other transformer-based method \cite{zhao2021point}.
	\item \textbf{Classification:} Following MaskPoint \cite{yu2022point}, we compare with MaskPoint \cite{yu2022point}, PointBERT\cite{pang2022masked}, Simple3D-Former \cite{wang2022can} and P2P \cite{wang2022p2p}.
\end{itemize}

\subsection{Detection}
\label{sec:exp3}
For many contemporary 3D pre-training works, they report the performance mainly on object-level classification and part segmentation tasks, which is insufficient in real-world point cloud tasks. Moreover, the most recent P2P \cite{wang2022p2p} needs to project the point cloud into 2D views, which is confronted with great challenges in solving real-world point cloud tasks. Our EPCL does not require any projections and thus can be widely applied to real-world point cloud scenarios. In this section, the detection on ScanNet V2 is evaluated and compared with the state-of-the-art 3D pre-training approaches \cite{pang2022masked,yu2022point} and object detection methods\cite{vu2022softgroup,wang2022cagroup3d}.

\begin{table}[ht]
	\centering
	\caption{Detection on ScanNet V2.}
	
	\begin{tabular}{c|ccc}
		\hline
		Method & 3D Pretrain & AP$_{50}$ & AP$_{25}$ \\ \hline
		PointBERT   & \Checkmark   & 38.3 & 61.0 \\
		MaskPoint & \Checkmark    & 42.1  & 64.2 \\ 
		TAP      & \Checkmark    & 41.4 & 63.0 \\ \hline
		
		Simple3D-Former & \XSolidBrush   & 40.7  & 59.4\\
		CLIP frozen + 3DETR      & \XSolidBrush  & \bf 43.0 & \bf 62.6\\ \hline
		SoftGroup & \XSolidBrush   & 59.4 & 71.6 \\
		CAGroup3D & \XSolidBrush   & 60.8  & 73.6\\
		CLIP frozen + CAGoup3D  & \XSolidBrush  & \bf 61.1 & \bf 73.7 \\ \hline
	\end{tabular}
	\label{tab:det}
\end{table}

Table \ref{tab:det} shows that the frozen CLIP model achieves better accuracy than baseline methods. This observation shows that the CLIP transformer can effectively learn 3D representation to solve real-world 3D detection and achieve better performance than state-of-the-art 3D pre-training method, TAP \cite{wang2023take}. Note that the CLIP model is frozen and has not seen any 3D point cloud in their learned parameters. Our method only fine-tunes the same training dataset with other baselines. These results demonstrate that the CLIP transformer achieves better accuracy. Notably, EPCL achieves better performance than the state-of-the-art object detection method CAGroup3D when using the head of CAGroup3D. The impressive performance is attributed to the strong ability of the CLIP model to align the features in different modalities.

\subsection{Semantic segmentation}
\label{sec:exp4}

\textbf{Indoor semantic segmentation.} This section introduces the experiments on indoor segmentation dataset S3DIS\cite{armeni20163d}. To ensure fair comparison, we put all these encoders on the same code base, which shares the same hierarchical tokenizer and semantic task head. The MaskPoint initializes the encoder with the pre-trained model of MaskPoint, and the Simple3D-Former initializes the encoder with 2D ViT. Then, the MaskPoint and Simple3D-Former methods finetune the \emph{whole model} on the S3DIS training samples. In contrast, our method keeps the CLIP model frozen and \emph{only} finetunes the tokenizer and task head.

\begin{table}[ht]
	\centering
	\caption{Indoor semantic segmentation on S3DIS (Area5).}
	
	\begin{tabular}{c|ccc}
		\hline
		Method &OA  & mAcc.  & mIoU.  \\ \hline
		Point Transformer & 90.8 &  76.5   & 70.4  \\ \hline
		MaskPoint & 89.0  &  73.8   & 67.1  \\  
		Simple3D-Former & - & 72.5    &  67.0 \\ \hline
		Ours                          &\bf 90.8     & \bf 77.8 & \bf 71.5 \\ \hline
	\end{tabular}
	\label{tab:indoor_sem_seg}
\end{table}

Table \ref{tab:indoor_sem_seg} shows that the frozen CLIP model obtains obviously better accuracy (\emph{i.e.}, mAcc. and mIoU) than other state-of-the-art 3D pre-training methods as well as Point Transformer on S3DIS (Area5) dataset. This observation illustrates that the frozen CLIP model is an efficient point cloud  learner in the real-world semantic segmentation task.

\textbf{Outdoor semantic segmentation.} We also evaluate our EPCL on outdoor segmentation task and compare with recent task-specific methods (Cylinder3D \cite{zhou2020cylinder3d}, PVKD \cite{hou2022point}, 2DPASS \cite{yan20222dpass}, RPVNet \cite{xu2021rpvnet}) and pretrained method (RangeFormer \cite{kong2023rethinking}). Quantitative comparison on SemanticKITTI validation set is shown in Table \ref{tab:outdoor_sem_seg}, which demonstrates better accuracy than these compared methods. It is worth noting that the frozen CLIP achieves the highly competitive performance compared to the task-specific model, even without any engineering tricks such as test time augmentation and model ensemble.

\begin{table}[ht]
	\centering
	\caption{Outdoor semantic segmentation on SemanticKITTI.}
	
	\begin{tabular}{c|ccc}
		\hline
		Method & mIoU. (val.)  \\ \hline
		
		Cylinder3D &  65.2 \\
		PVKD       &  66.4 \\
		2DPASS     &  69.3 \\
		RPVNet     &  69.6 \\ \hline
		RangeFormer &  69.6 \\
		Ours       & \bf{72.4} \\ \hline
	\end{tabular}
	\label{tab:outdoor_sem_seg}
\end{table}

\begin{table}[ht]
	\centering
	\caption{Classification on ModelNet40.}
	
	\begin{tabular}{c|cc}
		\hline
		Method  & 3D Pretrain &OA \\ \hline
		PointBERT    & \Checkmark   & 93.2  \\ 
		MaskPoint  & \Checkmark & \bf 93.8 \\ \hline
		P2P        & \XSolidBrush    & 92.7 \\
		Simple3D-Former & \XSolidBrush   & 92.0 \\ 
		Ours + w/o CLIP frozen  & \XSolidBrush   & 92.3 \\ 
		Ours                               & \XSolidBrush   & \bf 92.9 \\ \hline
	\end{tabular}
	
	\label{tab:cls}
\end{table}

\begin{table*}[ht]
	\begin{center}
		\caption{Few-shot learning accuracy of 3D pre-training methods and EPCL on ModelNet40.}	
		\begin{tabular}{c|ccccc}	
			\hline
			\textbf{Tuning Method} 
			&\textbf{5-w,10-s} 
			&\textbf{5-w,20-s} 
			&\textbf{10-w,10-s} 
			&\textbf{10-w,20-s}
			&\textbf{30-w,10-s} \\
			\hline
			PointBERT
			&94.6 $\pm$ {3.1}
			&96.3 $\pm$ 2.7
			&91.0 $\pm$ {5.4}
			&92.7 $\pm$ 5.1 
			&81.4 $\pm$ 2.4 \\
			
			MaskPoint
			&\underline{95.0} $\pm$ 3.7
			&\underline{97.2} $\pm$ \underline{1.7}
			&\textbf{91.4 $\pm$ 4.0 }
			&93.4 $\pm$ \textbf{3.5} 
			&80.7 $\pm$ 4.9 \\
			\hline

			Ours
			& \textbf{95.1 $\pm$ 2.7} & \textbf{97.3 $\pm$ 1.6} & \underline{91.1} $\pm$ \underline{4.2} & \textbf{93.5} $\pm$ \underline{3.8} & \textbf{81.7 $\pm$ 0.7} \\
			
			\hline	
			
		\end{tabular}
		\label{t_fewshot_main}
	\end{center}
\end{table*}
\subsection{Classification}
\label{sec:exp2}
\paragraph{Supervised classification.}Table \ref{tab:cls} summarizes the classification results on the synthetic ModelNet40 dataset. Our method achieves comparable performance to the state-of-the-art 3D pre-training methods although our method does not pre-train the model on the object dataset, \emph{i.e.}, ShapeNet. Compared to the P2P, which is the recent state-of-the-art method that directly used a 2D pre-trained model, our method achieves better accuracy. P2P designs a projection module to render images from 3D objects which is tailored for 3D classification. Our method does not need to project 3D to 2D images and directly processes the 3D tokens, which shows potential in other applications except classification. Compared to Simple3D-Former, which finetunes the whole model from a 2D pre-trained model, our method still achieves better accuracy. These positive results verify our argument that \emph{the frozen CLIP transformer is an effective encoder to learn 3D representation for point cloud understanding}.

\textbf{Few-shot learning.} One important advantage of pre-trained models is that fewer training samples are required in downstream tasks. This is usually evaluated by the few-shot learning task.
To evaluate the few-shot learning ability, we follow MaskPoint \cite{pang2022masked} to conduct experiments with the setting of "$K$-way $N$-shot", \emph{i.e.}, \textit{5way-10shot, 5way-20shot,10way-10shot and 10way-20shot}.

Table \ref{t_fewshot_main} summarizes the comparison experiments on "$K$-way $N$-shot" few-shot learning. Our EPCL obtains more accurate classification results than the state-of-the-art 3D pre-training methods. This observation shows that the frozen CLIP transformer is an effective representation learning encoder in the challenging few-shot learning setting. 

\subsection{Ablation studies and discussion}
\label{sec:exp5}
In this section, we introduce several key ablation studies to examine the effect of each component of our EPCL. More ablation studies and discussions can be found in the supplement.

\textbf{Task token.}  The task token module aims to learn task embedding for a specific task. As the task embedding module is only trainable in the training stage and the parameters and initialization are frozen during the inference stage. To demonstrate its effectiveness, we conduct an ablation study by removing the task token module on ScanNet V2 at the detection task. Table \ref{tab:ablation_1} shows that the detection accuracy decreases ($1.9\%$) when the task embedding is discarded. 
This experiment illustrates that learning additional task-related feature bias is beneficial to the CLIP model in point cloud tasks. 

\begin{table}[ht]
	\centering
	\caption{Ablation studies of the task embedding and the frozen strategy on ScanNet V2 detection task.}
	
	\begin{tabular}{cccc}
		\hline
		task token & CLIP Frozen  & AP$_{50}$ & Train Para. (\%) \\ \hline
		
		\XSolidBrush & \Checkmark &  59.2  & 55.23 \\
		\Checkmark  & \XSolidBrush &  60.1 & 100\\
		\Checkmark  & \Checkmark & \bf 61.1 & 55.51 \\ \hline
	\end{tabular}
	
	\label{tab:ablation_1}
\end{table}

\textbf{Task token transferability.} To demonstrate the transferability of our task token, we use the detection task token to replace the classification task token. The classification result improves by 5.7\% by using the detection task token compared to the random one. The improved performance of using the detection task token over the random task token demonstrates the transferability. Moreover, the accuracy will drop \textgreater 10\% when directly using other task tokens. This result shows that the task token needs to be fine-tuned together with the tokenizer. Simply replacing the task token will lead to inferior performance.

\textbf{CLIP Frozen or not?} Recall that we freeze the CLIP model during the entire training process. It is natural to wonder what the performance will be if turning the parameters of the CLIP model during training. To answer this question, we turn the whole neural network on during the training stage and conduct an ablation study on real-world detection task. Table \ref{tab:ablation_1} shows that the accuracy drops 1.0\% when the CLIP model is turned on. \emph{The reason} is that the relatively small-scale 3D training dataset fine-tunes the CLIP model to a worse parameter space compared to the one trained on the large-scale dataset.
Also, our method shows that the frozen CLIP transformer achieves better training efficiency than the version without freezing.

\begin{table}[ht]
	\centering
	\caption{Results of other 2D pre-trained models on the ScanNet V2 detection task.}
	
	\begin{tabular}{c|cc}
		\hline
		Method &  AP$_{50}$ & AP$_{25}$ \\ \hline
		SAM    & 59.5  & 73.7\\
		DINO   & 60.0  & 72.7\\ \hline
		
		Ours   & \bf 61.1 & \bf 73.7 \\ \hline
	\end{tabular}
	
	\label{tab:2Dpretrain}
\end{table}

\textbf{Is CLIP better than other 2D pretrained models?} The CLIP model is trained on a large-scale dataset that pairs internet images with text, containing a wide range of real-world multimodal knowledge. To demonstrate its effectiveness in 3D representation learning, we replace the frozen CLIP transformer with other 2D pretrained models (SAM \cite{kirillov2023segany}, DINO \cite{caron2021emerging}) that are solely trained on images (single modality). We then freeze the model during the training stage. Table \ref{tab:2Dpretrain} illustrates that CLIP achieves higher detection accuracy on ScanNet-V2 compared to other 2D pretrained models. This result demonstrates that the CLIP transformer, with its multimodal knowledge, outperforms 2D pretrained models that are only trained on images.

\section{Conclusion}
\label{sec:conclusion}
This paper proposes an efficient yet effective method to construct point cloud understanding models by using the frozen CLIP transformer. Our method converts the input point cloud into sequential tokens with a point tokenizer. These tokens and the learnable task token input into the frozen CLIP transformer can generate robust 3D representation.  We conduct thorough analyses of the inner mechanism and find the tokenizer can weakly align the 3D and 2D features at different modalities. Then, the CLIP transformer can align them further. Our method achieves appealing performance on a wide range of downstream tasks, including both real-world detection and segmentation tasks as well as synthetic object-level classification tasks.

{\small
	\bibliographystyle{aaai24}
	\bibliography{egbib}
}

\clearpage
\section{Appendices} 
 \subsection{More experiments}
 
 \textbf{Prompt or not?}
 In our experiments, we found that our task token is similar to prompt learning \cite{jia2022visual}. We also conducted experiments to compare with prompt learning. Following the VPT \cite{jia2022visual}, we added the task token to every transformer layer of Frozen CLIP. The result is shown in Table \ref{tab:prompt}, which shows that our task token obtains slightly worse results in classification but better in the real-world detection task. 
 
 \begin{table}[h]
 	\centering
 	\caption{Prompt learning for Frozen CLIP.}
 	\begin{tabular}{c|cc}
 		Method &  Cls. & Det. \\ \hline
 		Prompt    & \bf 93.03 & 42.43  \\
 		Ours (task token)      &  92.91 & \bf 43.00 \\ \hline
 	\end{tabular}
 	\label{tab:prompt}
 \end{table}
 
 The reason may be that the parameter space of Frozen CLIP is optimized well by huge image-text pairs. The learned prompts on a small 3D dataset may slightly undermine the Frozen CLIP. Using the task token only at the input of transformer will keep the integrity of Frozen CLIP. Moreover, adding more tokens in the transformer layers will result in more training and inference costs. Based on the above observations and considerations, we select the task token solution in our method.

 We have taken more experiments for the 3D classification task on ScanObjectNN. Table \ref{tab:objectNN} demonstrates better classification accuracy on real-world point clouds.
 
 \begin{table}[h]
 	\vspace{-0.3cm}
 	\centering
 	\caption{Classification Results of ScanObjectNN.}\label{tab:objectNN}
 	\begin{tabular}{l c c c}
 		\hline
 		Method & OBJ & BG & PB \\
 		PointNet    & 79.2  & 73.3  & 68.0\\
 		DGCNN   & 86.2  & 82.8  & 78.1 \\ \hline
 		PointViT-OcCo   & 85.5 & 84.9   & 78.8 \\
 		Ours    & 87.6 & 85.7 & 79.7 \\
 		\hline
 	\end{tabular} 
 \end{table}
 
 \textbf{Density variants.} The outdoor and indoor point clouds have large difference in density. To evaluate the robustness on density variants, we conduct experiments by downsamping the ScannetV2 to 50\% and 30\% of the original points. Then, the detection accuracy is evaluated by following the same setting with the main experiments. The detection accuracy is 42.1\% and 39.0\% which is still better than 3DETR(37.5\%). This demonstrates our method is robust to density variants.

\begin{figure*}[t]
 	\centering
 	\includegraphics[width=\linewidth]{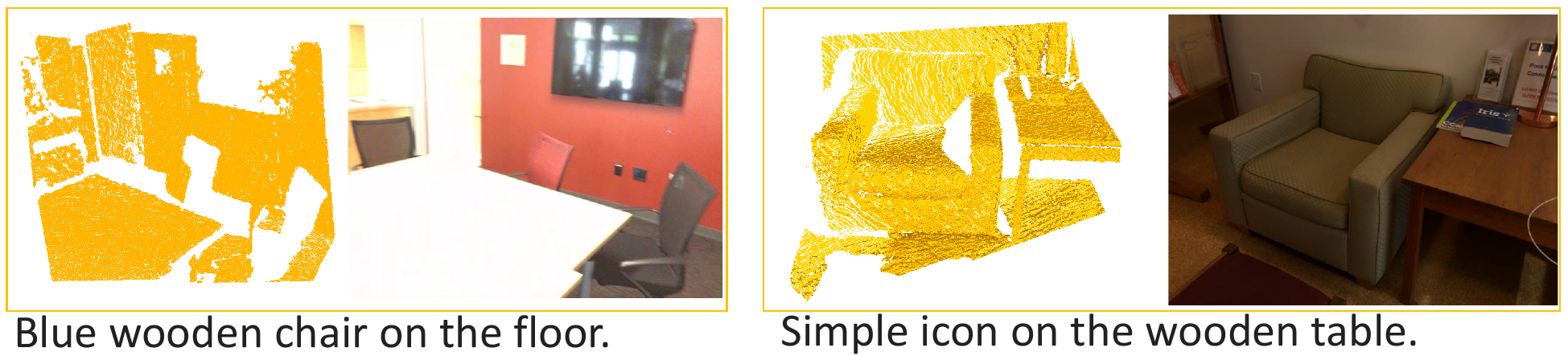}
 	\caption{Our method can generate text for 3D point cloud by concatenating with a caption head. The 2D image is the corresponding region for the point cloud. }
 	\label{fig:caption}
 \end{figure*}

 \textbf{Can frozen CLIP transformer generate text for 3D point cloud?} We use the trained model of detection task as the backbone and used it to replace the CLIP model in the CLIP-Caption method \cite{mokady2021clipcap} to generate text for the given 3D point cloud. Figure \ref{fig:caption} shows that our method can generate reasonable captions for 3D point clouds. 
 
 \subsection{Visual results}
 In this section, we show some visual results to demonstrate the quality of point cloud features learned by Frozen CLIP.
 
\begin{figure*}[p]
 	\centering
 	\includegraphics[width=\linewidth]{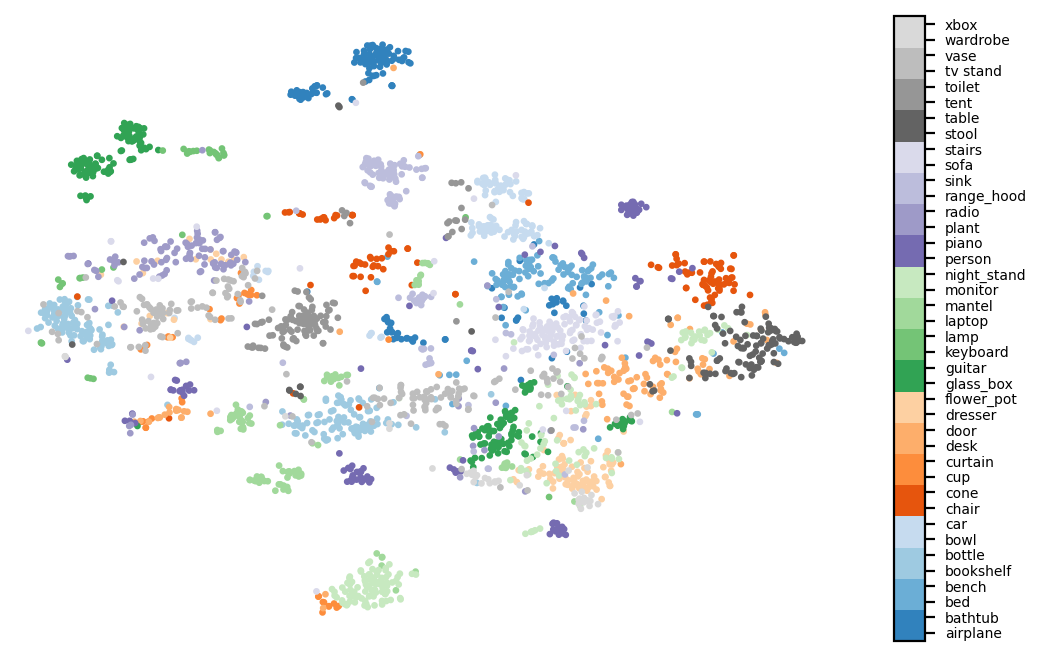}
 	\caption{The TSNE map of features for different categories in the ModelNet40 dataset.}
 	\label{fig:tsne40}
 \end{figure*}

 \textbf{Classification.} 
 The feature quality dominates the classification accuracy. To visually see the feature quality, we project the CAD object features of ModelNet40 into TSNE. Figure \ref{fig:tsne40} shows that the Frozen CLIP can cluster different categories into different positions. Each class can be classified at different positions.
 
\begin{figure}[H]
 	\centering
 	\includegraphics[width=\linewidth]{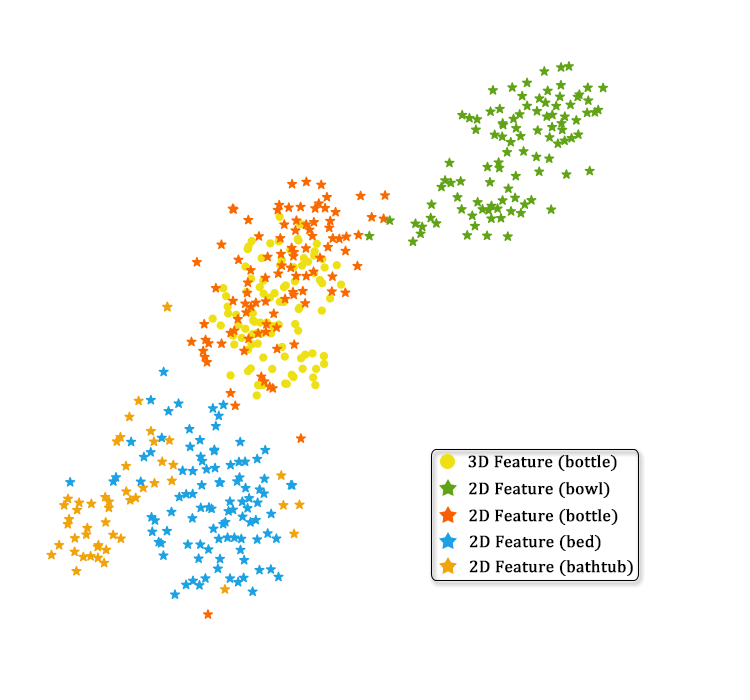}
 	\caption{The TSNE map of an example of 3D category (airplane) with different 2D image categories.}
 	\label{fig:tsne5}
 \end{figure}

 \textbf{Feature alignment.} One advantage of our EPCL is the feature alignment between 2D and 3D without paired data. Therefore, we can leverage multimodal benefits without paired data. Figure \ref{fig:tsne5} shows an example of 3D airplane and four other 2D modalities. The TSNE map illustrates that our EPCL can project the same category at different modalities into same position in the TSNE map. This observation demonstrates that our EPCL can align 3D and 2D at category level.
 
 \subsection{More implementation details of downstream tasks}
 The key settings of downstream tasks have been introduced in the main manuscript. In this section, we introduce more implementation details about optimization and hyper-parameters selection during our experiments. 
 
\begin{figure}[H]
 	\centering
 	\includegraphics[width=\linewidth]{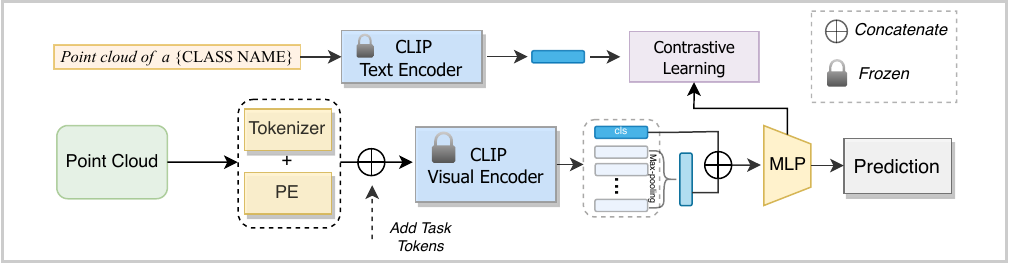}
 	\caption{The framework of classification.$PE$ means the position encoding.}
 	\label{fig:class}
 \end{figure}

 \textbf{Object classification.}
 For object classification, we follow the transformer-based methods \cite{yu2022point,pang2022masked} to utilize three-layer MLPs as the tokenizer. Besides the conventional cross-entropy loss which is applied to the classification head, deep features from the transformer are also compared with the CLIP text features of the label to optimize the tokenizer and task token. The contrastive loss function is the same as that in CLIP. The overall classification framework is shown in Figure \ref{fig:class}. In our experiments, we apply dropout(0.3) after the Frozen CLIP and dropout(0.2) for the first two layers of MLP module.
 
\begin{figure}[H]
 	\centering
 	\includegraphics[width=\linewidth]{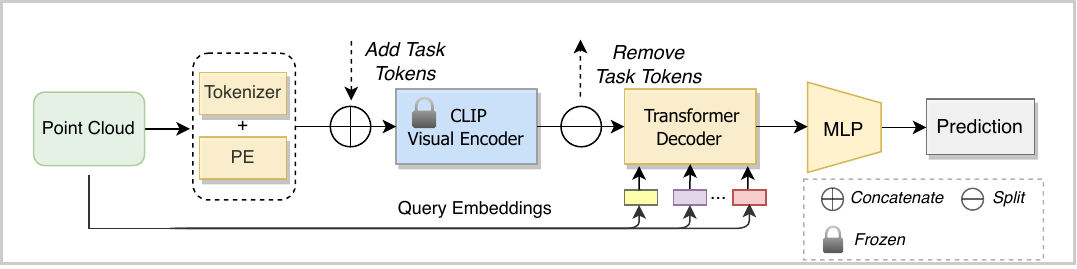}
 	\caption{The framework of detection using 3DETR as the detection head. 
 		\emph{Prediction} is the output of detection.}
 	\label{fig:3detr_detection}
 \end{figure}

\begin{figure}[H]
 	\centering
 	\includegraphics[width=\linewidth]{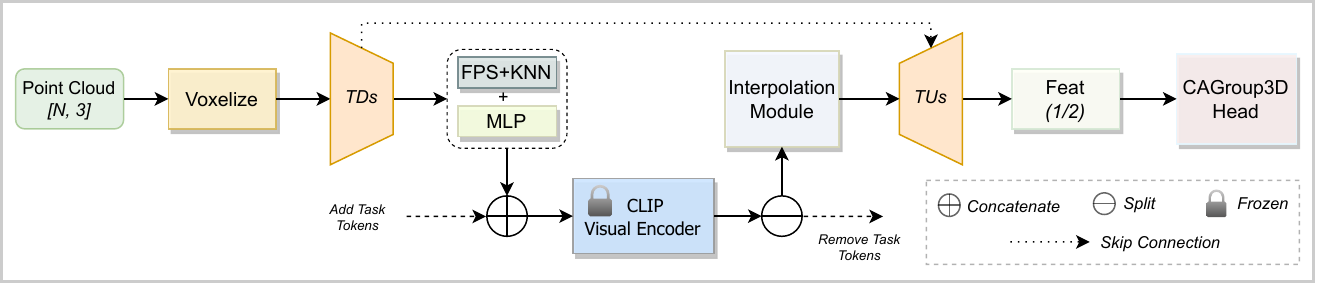}
 	\caption{The framework of detection using CAGroup3D as the detection head. $TD$ and $TU$ are defined for transition down and transition up respectively.}
 	\label{fig:cag3d_detection}
 \end{figure}

 \textbf{Indoor object detection.} Firstly,
 following MaskPoint \cite{liu2022masked}, we use the 3DETR \cite{misra2021end} as the detection head. The tokenizer and task embedding modules are the same as that for the classification task. The overall detection framework is shown in Figure \ref{fig:3detr_detection}. The final learning rate in our experiment is $5\times 10^{-7}$. Moreover, we apply dropout(0.3) after the Frozen CLIP. Secondly, we applied the frozen CLIP to a strong detection head, CAGroup3D \cite{wang2022cagroup3d}. Specifically, we apply three transition downsample and 3D sparse convolution residual blocks to convert the point cloud into the number of tokens varies per sample. Therefore, the farthest point sampling (FPS), K-nearest neighbour and MLP are applied to obtain fixed token numbers. 
 Those tokens are input into Frozen CLIP for deep feature extraction. Then, interpolation module are applied to restore the features corresponding to the downsampled 3D voxels. After that, we use two transition upsample to restore the features of half the resolution of the input point clouds, and concatenate the multi-scale features with the tokenizer features. These features input into CAGroup3D head for detection. The corresponding framework is shown in Figure \ref{fig:cag3d_detection}.
 
\begin{figure}[H]
 	\centering
 	\includegraphics[width=\linewidth]{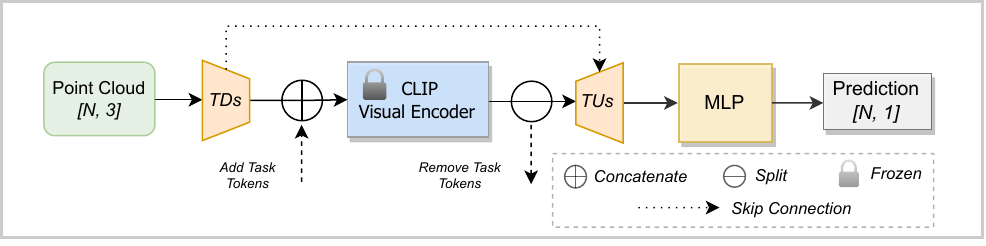}
 	\caption{The framework of indoor semantic segmentation. \emph{Prediction} is the output of semantic segmentation.}
 	\label{fig:indoor_segment}
 \end{figure}

 \textbf{Indoor semantic segmentation.} For the indoor semantic segmentation task, inspired by PointMixer \cite{choe2022pointmixer} and PointTransformer \cite{zhao2021point}, we apply four transition downsample and mixer blocks to convert the point cloud into tokens. After encoding the features of the downsampled tokens in the transformer, we gradually decode the point cloud back in the head at multiple scales and concatenate the multi-scale features with the tokenizer features. The overall indoor semantic segmentation framework is shown in Figure \ref{fig:indoor_segment}. We resort to the cross entropy loss during the finetuning.
 
\begin{figure}[H]
 	\centering
 	\includegraphics[width=\linewidth]{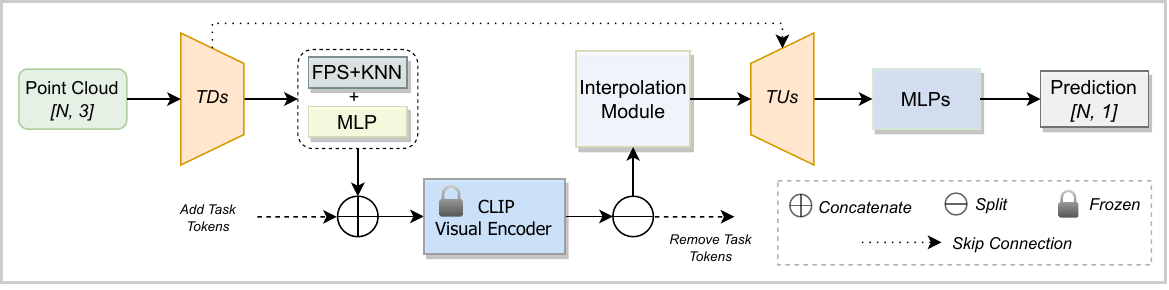}
 	\caption{The framework of outdoor semantic segmentation.}
 	\label{fig:outdoor_segment}
 \end{figure}

 \textbf{Outdoor semantic segmentation.} As shown in Figure \ref{fig:outdoor_segment}. For outdoor semantic segmentation, we adopted similar processing as CAGroup3D in the detection experiments. The difference is that here we perform 4 times downsampling on our tokenizer. After encoding the point cloud through the transformer, it is decoded back to the original input resolution size, and then fed into the segmentation head.
 
 To evaluate the few-shot learning ability, we conduct two experiments. One experiment follows the setting of "$K$-way $N$-shot"  \cite{pang2022masked,yu2022point}. It randomly selects $K$ classes and then samples $N$ samples on each class for training. The other experiment follows the setting of "16-shot" \cite{zhang2022pointclip} without pre-training. For the "$K$-way $N$-shot", it randomly selects $K$ classes and then samples $N$ samples on each class for training. For the "16-shot" few-shot learning, it randomly selects "16" samples for the whole 40 categories and uses the selected samples to train the neural network. This aims to evaluate the few-shot ability without seeing plenty of 3D objects. Then, 20 samples are picked for each class from the test set to evaluate the performance. These few-shot learning tasks are evaluated on ModelNet40. The previous studies adopt four evaluation settings, \emph{i.e.}, \textit{5way-10shot, 5way-20shot,10way-10shot and 10way-20shot}. To further evaluate the robustness of few-shot learning, we add a challenging setting, \emph{i.e.}, \emph{30-way,10-shot} and \emph{16-shot}.  The comparison methods are pre-trained on the ShapeNet. In comparison, our method does not require the pre-training step in few-shot learning.
 
 \begin{table}[H]
 	\centering
 	\caption{16-shot classification results on ModelNet40.}
 	
 	\begin{tabular}{c|c}
 		\hline
 		Method &  Acc (16-shot) \\ \hline
 		PointCLIP   & 83.8  \\
 		Simple3D-Former   & 76.2  \\ \hline
 		
 		Ours                                 & \bf 85.2 \\ \hline
 	\end{tabular}
 	
 	\label{tab:16shot}
 \end{table}
 
 Table \ref{tab:16shot} shows the comparative experiments on the 16-shot classification task. Our method achieves the best accuracy compared to the recent existing state-of-the-art few-shot learning methods. These experiments show that the 2D frozen CLIP transformer works well on limited training samples in the 3D point cloud classification task.

\end{document}